\documentclass{article}

\PassOptionsToPackage{numbers,square,sort&compress}{natbib}

 \usepackage[main, final]{neurips_2026}

\usepackage[utf8]{inputenc} 
\usepackage[T1]{fontenc}    
\usepackage{hyperref}       
\usepackage{url}            
\usepackage{booktabs}       
\usepackage{amsmath}        
\usepackage{amsfonts}       
\usepackage{nicefrac}       
\usepackage{microtype}      
\usepackage{xcolor}         
\usepackage[table]{xcolor}
\usepackage{graphicx}
\usepackage{wrapfig}
\usepackage{multirow}  
\usepackage{subfig}
\usepackage{float}
\usepackage{amsthm}
\usepackage{fontawesome5}
\newtheorem{proposition}{Proposition}[section]

\title{Not All Tasks Quantize Equally: Fisher-Guided Quantization for Visual Geometry Transformer}

\author{%
   \textbf{Yipu Zhang}$^{1,\dagger}$ \quad 
   \textbf{Jintao Cheng}$^{1,\dagger}$ \quad
   \textbf{Weilun Feng}$^{2,3,\dagger}$ \quad
   \textbf{Jiehao Luo}$^{4}$ \\
   \textbf{Chuanguang Yang}$^{2}$ \quad
   \textbf{Zhulin An}$^{2}$ \quad
   \textbf{Yongjun Xu}$^{2,5}$ \quad
   \textbf{Wei Zhang}$^{1,*}$ \\
   \\
   $^{1}$Department of Electronic and Computer Engineering, HKUST \\
   $^{2}$State Key Laboratory of AI Safety, Institute of Computing Technology,\\
   Chinese Academy of Sciences \\
   $^{3}$University of Chinese Academy of Sciences \\
   $^{4}$School of Electronic Science and Engineering (School of Microelectronics),\\
   South China Normal University \\
   $^{5}$Xiamen Institute of Data Intelligence \\
   \faGithub \,\textbf{Code:} \texttt{https://github.com/ypzhng/FGQ}
}

\begin{document}
\maketitle

\renewcommand{\thefootnote}{\fnsymbol{footnote}}
\footnotetext[2]{Equal contribution.}
\footnotetext[1]{Corresponding author:Wei Zhang<\texttt{wei.zhang@ust.hk}>.}
\renewcommand{\thefootnote}{\arabic{footnote}}
\setcounter{footnote}{0}

\begin{figure}[h]
    \centering
    \includegraphics[width=\linewidth]{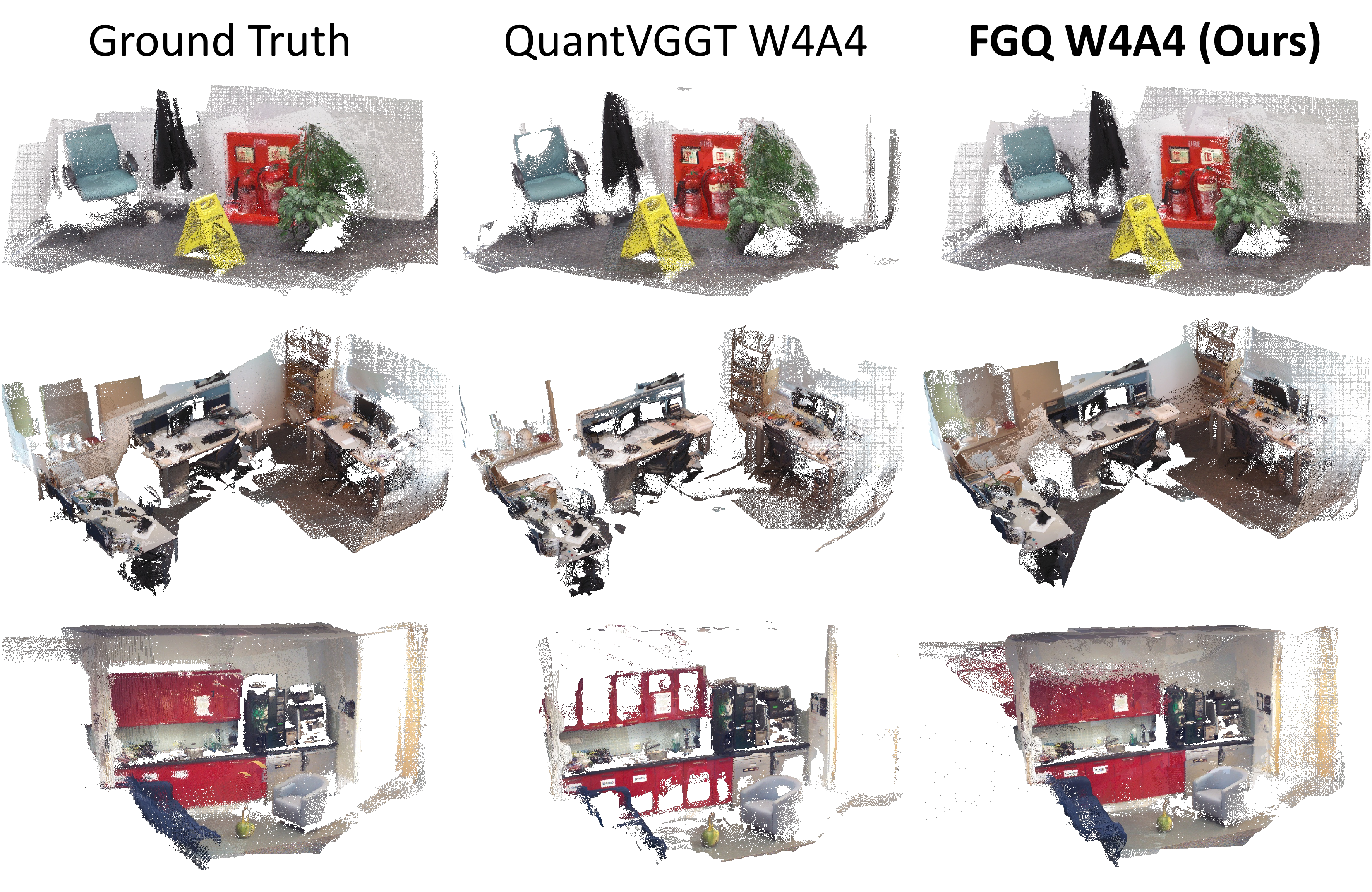}
    \caption{\textbf{FGQ} effectively preserves the point map reconstruction fidelity under 4-bit quantization, surpassing previous state-of-the-art VGGT (QuantVGGT~\citep{feng2025quantized}) quantization.}
    \label{fig:visual}
\end{figure}

\begin{abstract}
  Feed-forward 3D reconstruction models, represented by Visual Geometry Grounded Transformer (VGGT), jointly predict multiple visual geometry tasks such as depth estimation, camera pose prediction, and point map reconstruction in a single forward pass. They have been widely adopted in 3D vision applications, but their billion-scale parameters bring substantial memory and computation overhead, posing challenges for on-device deployment. Post-Training Quantization (PTQ) is an effective technique to reduce this overhead. Existing PTQ methods for feed-forward 3D models mainly focus on handling heavy-tailed activation distributions and constructing diverse calibration datasets. However, we observe that feed-forward 3D models predict multiple geometric attributes through a shared backbone, where different transformer blocks and hidden channels contribute distinctly to each task, resulting in substantially different sensitivities to quantization errors across tasks, blocks, and channels. Consequently, treating all tasks equally over-emphasizes insensitive tasks and causes significant accuracy loss on the sensitive ones. To address this issue, we propose Fisher-Guided Quantization (\textbf{FGQ}) for feed-forward 3D reconstruction models. Specifically, \textbf{FGQ} uses the diagonal Fisher information matrix to quantify the different sensitivities across tasks, blocks, and channels, and incorporates these sensitivities into the Learnable Affine Transformation during calibration to better preserve the channels and blocks most critical to each task. Extensive experiments across camera pose estimation, point map reconstruction, and depth estimation show that \textbf{FGQ} consistently outperforms state-of-the-art quantization baselines on VGGT, achieving up to 39\% relative improvement under the 4-bit quantization. 
\end{abstract}

\section{Introduction}


Feed-forward 3D reconstruction methods, represented by Visual Geometry Grounded Transformer (VGGT)~\citep{wang2025vggt}, have achieved significant progress by jointly predicting depth, camera poses, and point maps in a single forward pass. They have been widely adopted in downstream applications such as spatial reasoning~\citep{hu2025g, shen2026fine}, robotics~\citep{yang2026robo3r, liu2026palm}, and autonomous driving~\citep{lin2025vgd, cheng2024mf}. However, their billion-scale parameters introduce prohibitive computation and memory overhead during inference, limiting real-time processing and on-device deployment~\citep{zhang2026versaq}.

A variety of methods have been proposed to address this efficiency bottleneck, including quantization~\citep{feng2025quantized, zhang2026versaq, pan2026tail}, token merging~\citep{shen2025fastvggt, shu2025litevggt}, and architectural modifications that redesign the attention or computation flow for better efficiency~\citep{wang2025flashvggt, sun2025avggt}.

Existing PTQ methods for feed-forward 3D models primarily address two challenges: (i) mitigating activation outliers associated with special tokens (e.g., camera and register tokens), and (ii) capturing the geometric and semantic complexity of 3D representations during calibration dataset construction. However, these approaches overlook a critical characteristic of these models: their inherent multi-task architecture. Since depth, pose, and point map prediction operate at vastly different scales and are evaluated with distinct metrics, they exhibit markedly different sensitivities to quantization-induced perturbations.
\begin{wrapfigure}[16]{r}{0.4\textwidth}
    \centering
    \includegraphics[width=0.38\textwidth]{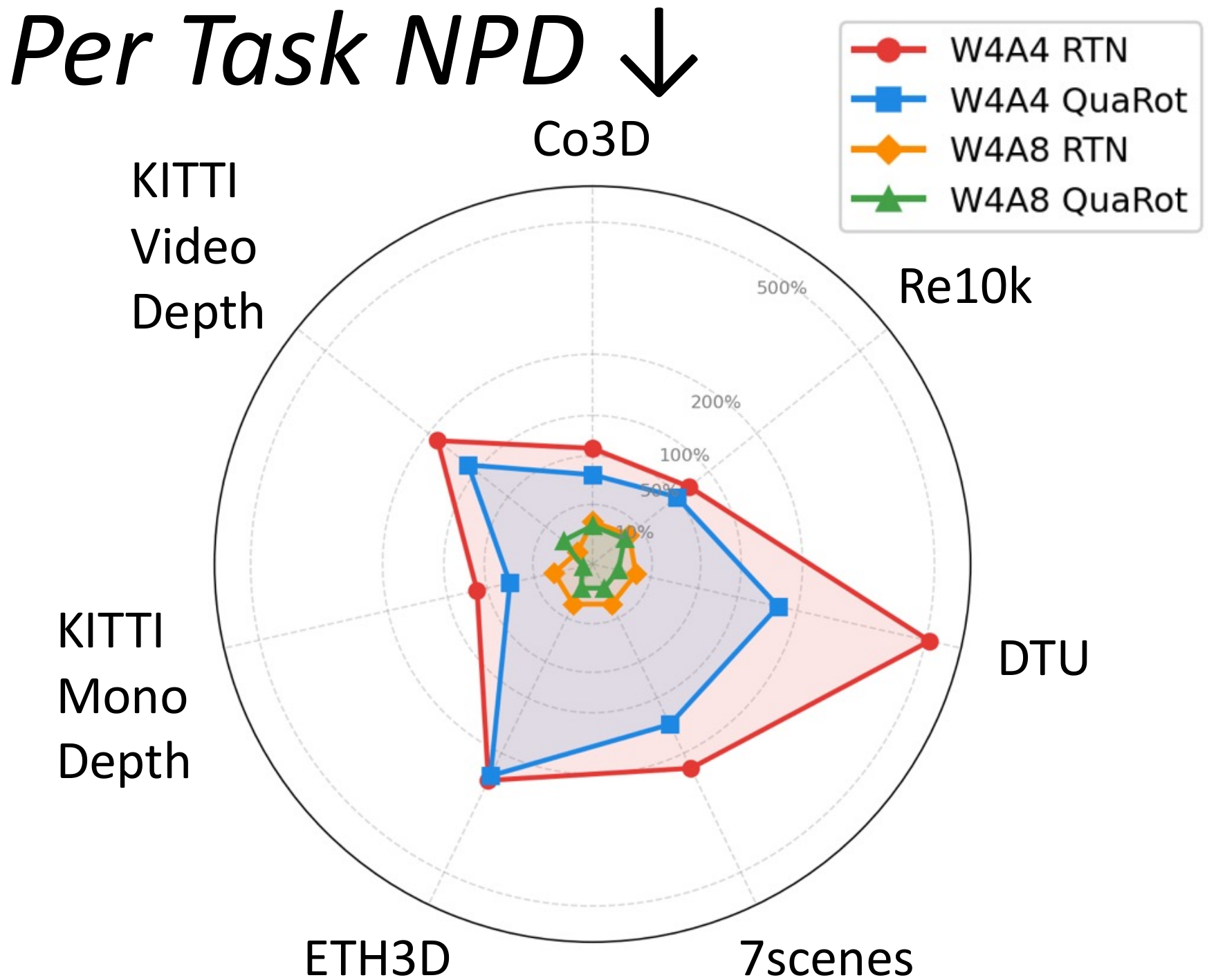}
    \caption{Normalized Performance Degradation (NPD) across various tasks (lower is better).}
    \label{fig:radar}
\end{wrapfigure}%
As illustrated in Fig.~\ref{fig:radar}, applying a uniform quantization strategy causes disproportionately severe degradation in point map reconstruction compared to camera pose and depth estimation. Beyond this task-level disparity, Fig.~\ref{fig:motivation} (a)\&(b) further shows that the per-block and per-channel quantization loss follows markedly different patterns across tasks. Together, these observations confirm that different transformer blocks and hidden channels contribute distinctly to each task, leading to substantially different quantization sensitivities across tasks, blocks, and channels.

To this end, we propose Fisher-Guided Quantization (\textbf{FGQ}), a task-aware PTQ framework for feed-forward 3D reconstruction models. \textbf{FGQ} leverages the diagonal Fisher information matrix to jointly quantify quantization sensitivity across three granularities: tasks (depth, pose, and point map), transformer blocks, and hidden channels. These sensitivities are then incorporated into the Learnable Affine Transformation during calibration, guiding the transformation to preserve the channels and blocks most critical to each task.


Our contributions are summarized as follows:
\begin{itemize}
  \item We identify the multi-task sensitivity problem in PTQ for feed-forward 3D reconstruction, showing that different transformer blocks and hidden channels contribute distinctly to each task, leading to substantially different quantization sensitivities across tasks, blocks, and channels.
  \item We propose Fisher-Guided Quantization (\textbf{FGQ}), which uses the diagonal Fisher information matrix to quantify per-task, per-block, and per-channel sensitivity, and integrates these sensitivities into the Learnable Affine Transformation objective to guide calibration toward the most quantization-sensitive components.
  \item Extensive experiments on camera pose estimation, point map reconstruction, and depth estimation demonstrate that \textbf{FGQ} improves task performance by up to 39\% over state-of-the-art PTQ baselines under 4-bit quantization, with the largest gains on point map reconstruction, where prior PTQ methods incur the most significant degradation.
\end{itemize}


\section{Related Work}


\paragraph{Feed-forward 3D reconstruction.}
Recent methods, including VGGT and $\pi^3$~\citep{wang2025vggt, wang2025pi}, directly predict scene geometry from images. Their learned representations support downstream rendering and scene understanding~\citep{hu2025g, yu2025core3d}. Related work on 3D reasoning and generation~\citep{yu2026elsa3d, shen2026fine}, further highlights the importance of connecting geometry with semantics. These applications motivate efficient 3D models. We therefore study VGGT quantization to reduce memory use while preserving reconstruction accuracy.

\paragraph{Efficient Feed-Forward 3D Reconstruction.}

Recent work targets the high compute and memory cost of feed-forward 3D reconstruction models such as VGGT. Token merging methods like FastVGGT~\citep{shen2025fastvggt} and LiteVGGT~\citep{shu2025litevggt} fuse similar tokens by exploiting redundancy in global attention. Architectural redesigns restructure attention itself: FlashVGGT~\citep{wang2025flashvggt} approximates global attention as cross-attention to a compact set of descriptor tokens generated by spatial resampling, while AVGGT~\citep{sun2025avggt} converts early global layers into frame attention and subsamples K/V over patch tokens. Closest to our work, recent PTQ methods~\citep{feng2025quantized, zhang2026versaq} address two key challenges in quantizing these models: (i) mitigating activation outliers from data-independent special tokens (e.g., camera and register tokens), which yield heavy tails and extreme channel and token variance unfriendly to low-bit quantization; and (ii) dealing with calibration data that captures the geometric and semantic complexity of 3D multi-view inputs. However, prior PTQ methods apply uniform quantization across all heads and treat feed-forward 3D models as single-task models, overlooking the distinct quantization sensitivities of their depth, pose, and point map predictions.

\paragraph{Model Quantization.}

Post-training quantization (PTQ) reduces the memory and compute cost of large language models. Weight-only methods such as GPTQ~\citep{frantar2022gptq} and AWQ~\citep{lin2024awq} compress weights to low bit-widths with minimal accuracy loss. To further quantize activations and the KV cache, a recent line of work applies affine transformations to suppress outliers before quantization. QuaRot~\citep{ashkboos2024quarot} uses fixed random Hadamard rotations, while SpinQuant~\citep{liu2024spinquant} and FlatQuant~\citep{sun2025flatquant} instead learn the transformation matrices during calibration: SpinQuant optimizes rotation matrices via Cayley optimization, and FlatQuant learns lightweight per-layer affine transformations that flatten weight and activation distributions. To better involve the sensitivity in the quantization process, we adopt the learnable affine transform based method as detailed in Sec.~\ref{sec:objective}.

\section{Method}
\subsection{Prelinminary}

Visual Geometry Grounded Transformer (VGGT)~\citep{wang2025vggt} is a feed-forward transformer that maps a set of $N$ input images $(I_i)_{i=1}^N$ to their 3D attributes (camera parameters, depth maps, point maps, and tracking features) in a single pass. Each image $I_i$ is first patchified by a DINOv2~\citep{oquab2023dinov2} encoder into $K$ tokens $t_i \in \mathbb{R}^{K \times C}$, and the full token set is denoted $\mathbf{T} = [t_1; t_2; \ldots; t_N] \in \mathbb{R}^{N \times K \times C}$. The backbone consists of $L$ \emph{Alternating-Attention} (AA) blocks, each interleaving two self-attention operations:
\begin{equation}
    \mathbf{T}_i^{(\ell + \tfrac{1}{2})} = \mathrm{FrameAttention}\big(\mathbf{T}_i^{(\ell)}\big), \quad i = 1, \ldots, N,
\end{equation}
\begin{equation}
    \mathbf{T}^{(\ell + 1)} = \mathrm{GlobalAttention}\big(\mathbf{T}^{(\ell + \tfrac{1}{2})}\big),
\end{equation}
where $\ell \in \{0, 1, \ldots, L-1\}$ indexes the AA block, 
$\mathbf{T}^{(\ell)}$ denotes the token state at the input of the 
$\ell$-th block (with $\mathbf{T}^{(0)} = \mathbf{T}$), and 
$\mathbf{T}^{(\ell + \tfrac{1}{2})}$ denotes the intermediate state 
after frame attention but before global attention within the same block. 
Here, $\mathrm{FrameAttention}(\cdot)$ operates independently within each 
image's $K$ tokens to preserve per-view structure, and 
$\mathrm{GlobalAttention}(\cdot)$ operates jointly over all $N \times K$ 
tokens to propagate information across views. The refined tokens $\mathbf{T}^{(L)}$ are then 
decoded by task-specific heads into the final 3D predictions.

\subsection{Fisher-Guided Quantization}

\begin{figure}
    \centering
    \includegraphics[width=\linewidth]{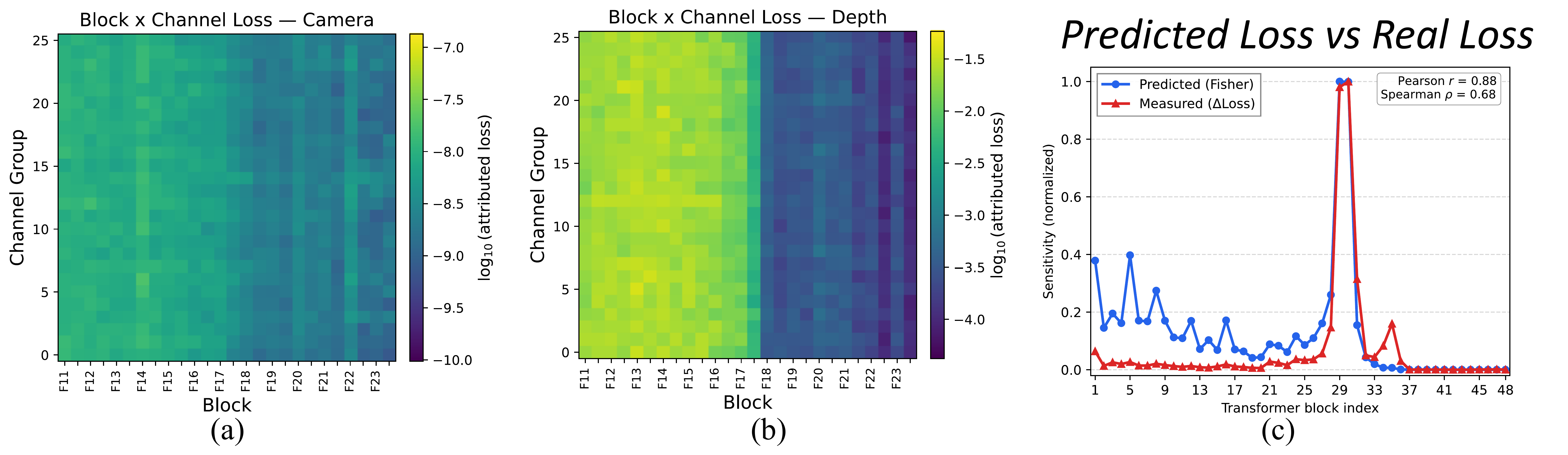}
    \caption{Fisher information accurately predicts quantization sensitivity and reveals task-specific patterns. (a) Per-block and per-channel quantization loss for camera pose estimation. (b) Per-block and per-channel quantization loss for depth estimation. The markedly quantization loss profiles between tasks motivate task-aware quantization strategies. (c) Fisher-predicted loss versus empirically measured loss across transformer blocks under W4A4 quantization, showing strong correlation ($r=0.88$). }
    \label{fig:motivation}

\end{figure}

\subsubsection{Sensitivity Varies Across Tasks}

To quantify the impact of quantization on each downstream task, we define
the Normalized Performance Degradation (NPD) of task $k$ at bit-width $b$ as
\begin{equation}
\mathrm{NPD}_k(b) \;=\; 
\frac{\bigl|\mathrm{Metric}_k(b)-\mathrm{Metric}_k(\mathrm{FP16})\bigr|}
     {\mathrm{Metric}_k(\mathrm{FP16})}\times 100\%.
\end{equation}

\paragraph{Task-level sensitivity.} As shown in Fig.~\ref{fig:radar}, applying W4A4 quantization to VGGT with PTQ methods leads to highly unbalanced degradation across tasks: camera pose estimation remains relatively robust, while point map reconstruction collapses with an NPD of $200\%\!\sim\!500\%$. This indicates that VGGT's shared backbone routes different geometric attributes through the same parameters, yet each task tolerates quantization noise to varying degrees. 

\paragraph{Block and channel-level sensitivity.} To locate the source of this asymmetry, we measure the per-block and per-channel quantization loss separately for the camera and depth tasks. As shown in Fig.~\ref{fig:motivation}(a) and (b), the two tasks produce clearly different loss profiles along both dimensions: blocks and channels that are critical for one task are often unimportant for the other. As a result, a uniform quantization policy is forced to compromise between tasks, and any objective that weights all tasks equally is dominated by the insensitive ones and therefore sacrifices accuracy on the sensitive ones. These observations motivate a task-aware design that allocates precision according to the sensitivity of each task at both the block and channel level.

\subsubsection{Fisher Information for Sensitivity Measurement}
\label{sec:fisher}
To realize this design, we introduce a task-aware sensitivity score at the block and channel level, which is cheap to compute and can be directly plugged into the calibration objective. We derive the score from a second-order expansion of each task loss, approximate the Hessian with the Fisher information to avoid its prohibitive cost, and further reduce it to its diagonal form for efficiency.

\begin{figure}
    \centering
    \includegraphics[width=\linewidth]{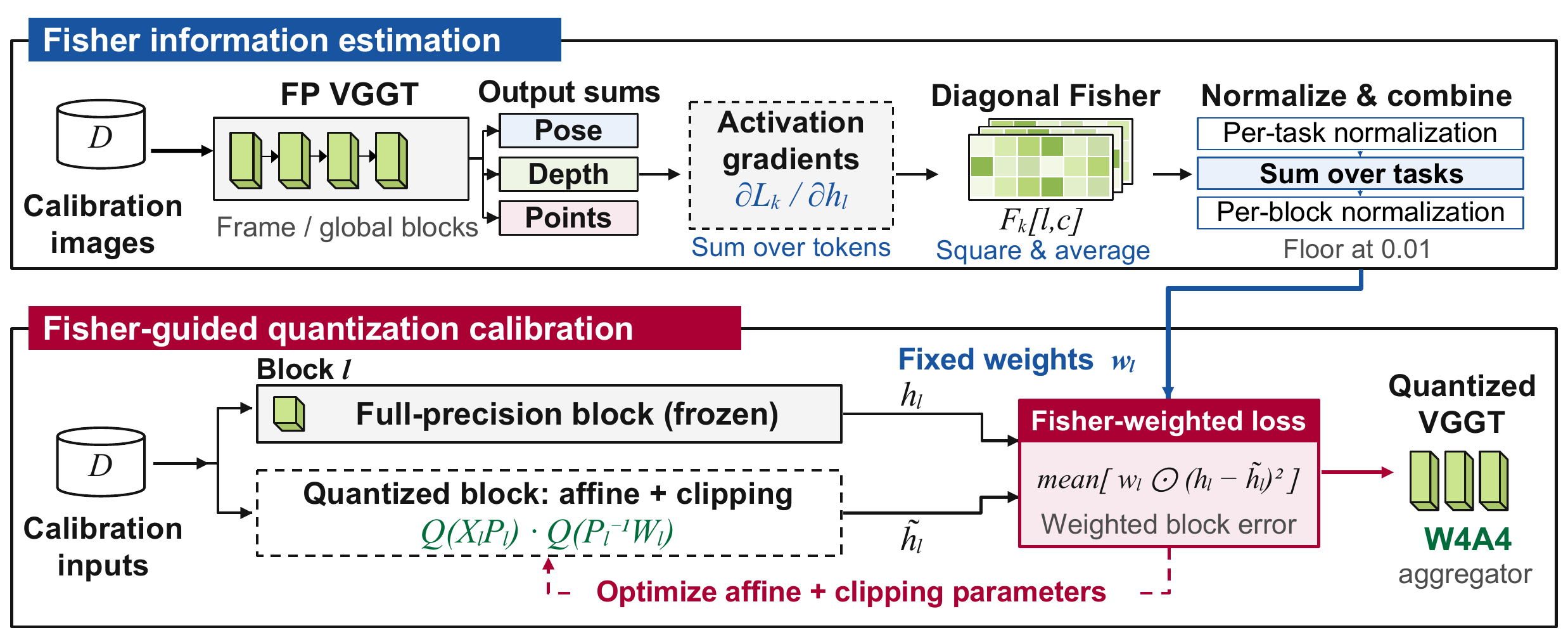}
    \caption{Overview of proposed Fisher-Guided Quantization (\textbf{FGQ}). We estimate channel-wise Fisher information from multiple geometry tasks (top) and use the resulting importance weights in the block-wise reconstruction loss to optimize affine and clipping parameters during VGGT quantization calibration (bottom).}
    \label{fig:overview}
\end{figure}

\paragraph{Quantization error and the role of the Hessian.}
Let $\mathbf{h}_l \in \mathbb{R}^{C}$ denote the output of block $l$,
where $C$ is the channel dimension, and let $\mathcal{L}_k$ denote
the loss of task $k$. Quantization perturbs $\mathbf{h}_l$ by
$\Delta \mathbf{h}_l$, which aggregates the effect of all weight and
activation rounding inside the block. A second-order Taylor expansion
of $\mathcal{L}_k$ around the full-precision output gives
\begin{equation}
\Delta \mathcal{L}_k \;\approx\;
\mathbf{g}_k^{\top}\Delta \mathbf{h}_l
\;+\;\tfrac{1}{2}\,\Delta \mathbf{h}_l^{\top}\mathbf{H}_k\,\Delta \mathbf{h}_l,
\end{equation}
where $\mathbf{g}_k$ and $\mathbf{H}_k$ are the gradient and Hessian
of $\mathcal{L}_k$ with respect to $\mathbf{h}_l$. A pre-trained VGGT
sits near a local optimum, and therefore
$\mathbb{E}[\mathbf{g}_k]\!\to\!\mathbf{0}$ and the first-order term
vanishes in expectation~\citep{lecun1989optimal}. The expected loss
change is governed by the Hessian term,
\begin{equation}
\mathbb{E}[\Delta \mathcal{L}_k]
\;\approx\;\tfrac{1}{2}\,\Delta \mathbf{h}_l^{\top}\mathbf{H}_k\,\Delta \mathbf{h}_l,
\label{eq:hessian_quad}
\end{equation}
which assigns a task-specific cost to each direction of perturbation:
directions with large eigenvalues amplify quantization error
into task loss, while directions with small eigenvalues absorb it.

\paragraph{Fisher information as a tractable surrogate.}

While Eq.~\ref{eq:hessian_quad} formalizes the appropriate notion of
block-wise sensitivity, direct evaluation of \(\mathbf{H}_k\) is
computationally prohibitive in our setting. Second-order automatic
differentiation incurs substantial runtime overhead per block and per
task, and explicit storage of \(\mathbf{H}_k\) requires
\(\mathcal{O}(C^2)\) memory per block. To avoid these costs, we use a
Fisher-based surrogate for the Hessian. This substitution is common in
the pruning and quantization literature~\citep{kwon2022fast,yang2024fisher}
and is motivated by the following standard identity.

For our calibration setting, \(x\) denotes a set of input images. 
The variable \(y\) denotes the target signal used
to compute the task loss \(\mathcal{L}_k\). We write
\(q_k(x,y)=q_k(x)q_k(y\mid x)\) for the task-specific calibration
distribution. For readability, the proposition below omits the task
subscript and writes \(q\) instead of \(q_k\).

\begin{proposition}[Conditional Hessian--Fisher identity]
\label{prop:fisher_hessian}
Let \(z\) be a differentiable argument of a normalized conditional
model \(p_z(y\mid x)\), where \(z\) may denote model parameters or an
intermediate activation treated as a local input to the remaining
network. Define the negative log-likelihood loss $
\ell(z;x,y)=-\log p_z(y\mid x)$.
Assume standard regularity conditions that allow differentiation under
the integral. If, at \(z^\star\), the data conditional distribution is
matched by the model, i.e.,
$q(y\mid x)=p_{z^\star}(y\mid x)$
for \(q(x)\)-almost every \(x\), then
\begin{equation}
\mathbb{E}_{x\sim q(x),\,y\sim q(y\mid x)}
\!\left[\nabla_z^2 \ell(z^\star;x,y)\right]
=
\mathbb{E}_{x\sim q(x),\,y\sim q(y\mid x)}
\!\left[
\nabla_z \ell(z^\star;x,y)
\nabla_z \ell(z^\star;x,y)^{\!\top}
\right].
\label{eq:hessian_fisher_identity}
\end{equation}
The matrix on the right-hand side is the Fisher information matrix with
respect to \(z\).
\end{proposition}

We give a self-contained proof of
Proposition~\ref{prop:fisher_hessian}, which is an instance of the
second Bartlett identity, in Appendix~\ref{app:proof}. The proposition
shows that, under a correctly specified likelihood and population
expectation, the expected Hessian of the negative log-likelihood equals
the Fisher information matrix. In practice, we use this identity as a
motivation for replacing the block Hessian by an empirical Fisher
estimated from first-order gradients at a well-trained checkpoint.

Instantiating this surrogate at the output of block \(l\) gives
\begin{equation}
\mathbb{E}[\Delta \mathcal{L}_k]
\;\approx\;
\tfrac{1}{2}\,
\Delta \mathbf{h}_l^{\top}
\mathbf{F}_k^{h_l}
\Delta \mathbf{h}_l,
\qquad
\bigl[\mathbf{F}_k^{h_l}\bigr]_{c,c'}
=
\mathbb{E}_{(x,y)\sim\mathcal{D}_{\mathrm{cal}}}
\!\left[
\frac{\partial \mathcal{L}_k(x,y)}{\partial h_{l,c}}\,
\frac{\partial \mathcal{L}_k(x,y)}{\partial h_{l,c'}}
\right].
\label{eq:fisher_quad}
\end{equation}
Here \(\mathcal{D}_{\mathrm{cal}}\) is a small calibration set. Since
\(\mathbf{F}_k^{h_l}\) only uses first-order gradients, it can be
estimated with one backward pass per task. Moreover, evaluating the
Fisher at the block output is convenient because
\(\Delta \mathbf{h}_l\) already contains the total perturbation induced
by quantizing block \(l\). Thus a single block-level Fisher term captures
the joint sensitivity of all quantized operations inside the block. As
shown in Fig.~\ref{fig:motivation}(c), this Fisher-based score is also
empirically predictive of the measured quantization loss, with Pearson
correlation \(r=0.88\).

\paragraph{Diagonal approximation.}
Although the Fisher surrogate avoids second-order derivatives, the full
matrix \(\mathbf{F}_k^{h_l}\) is still a \(C \times C\) matrix for each
block and task. This is costly to store and its off-diagonal entries can
be hard to estimate from a small calibration set. We therefore use the
diagonal empirical Fisher:
\begin{equation}
F_k[l,c]
=
\mathbb{E}_{(x,y)\sim\mathcal{D}_{\mathrm{cal}}}
\left[
\left(
\frac{\partial \mathcal{L}_k(x,y)}{\partial h_{l,c}(x)}
\right)^2
\right]
\approx
\frac{1}{N}
\sum_{n=1}^{N}
\left(
\frac{\partial \mathcal{L}_k(x_n,y_n)}
{\partial h_{l,c}(x_n)}
\right)^2 .
\label{eq:fisher_diag_def}
\end{equation}
This reduces storage from \(\mathcal{O}(C^2)\) to \(\mathcal{O}(C)\).
The quantity \(F_k[l,c]\) is the second moment of the per-sample
gradient on channel \(c\). Under the population Fisher identity, where
the expected score is zero, it also equals the gradient variance.
Keeping only the diagonal discards cross-channel terms and treats
channel perturbations as locally independent. This approximation is
sufficient for our goal, since we use the score to rank channel
importance rather than to recover the full curvature matrix.

With this approximation, Eq.~\ref{eq:fisher_quad} becomes
\begin{equation}
\mathbb{E}[\Delta \mathcal{L}_k]
\approx
\frac{1}{2}
\mathbb{E}_{x\sim\mathcal{D}_{\mathrm{cal}}}
\left[
\sum_{c=1}^{C}
F_k[l,c]\,
\bigl(\Delta h_{l,c}(x)\bigr)^2
\right].
\label{eq:fisher_diag}
\end{equation}
Thus, the loss increase is approximated by a channel-wise weighted
reconstruction error at the block output. Each \(F_k[l,c]\) measures how
sensitive task \(k\) is to squared error on channel \(c\) of block
\(l\). Compared with a uniform reconstruction loss, this form assigns
larger weights to channels that are more important for the task. We use
these task-aware channel weights in the learnable affine transformation
described next.

\subsubsection{Integrating Fisher Information into Learnable Affine Transformations}
\label{sec:objective}

Building on the per-channel Fisher score \(F_k[l,c]\) from
Section~\ref{sec:fisher}, we use task sensitivity to guide the calibration of
learnable affine transformations. Let \(l \in \{1,\ldots,L\}\) index the
calibrated transformer blocks, \(c \in \{1,\ldots,C\}\) index hidden channels,
and \(k \in \{1,\ldots,K\}\) index VGGT task heads.

\paragraph{Block-wise calibration objective.}
Learnable affine transformation methods~\citep{liu2024spinquant,sun2025flatquant}
insert an invertible matrix \(\mathbf{P}_l\) into a linear layer and fold its
inverse into the weight:
\begin{equation}
\mathbf{Y}_l
=
\bigl(\mathbf{X}_l \mathbf{P}_l\bigr)
\bigl(\mathbf{P}_l^{-1}\mathbf{W}_l\bigr)
=
\mathbf{X}_l\mathbf{W}_l .
\end{equation}
Quantization is then applied to the transformed activation
\(\mathbf{X}_l\mathbf{P}_l\) and transformed weight
\(\mathbf{P}_l^{-1}\mathbf{W}_l\), whose distributions are easier to quantize
while preserving the full-precision computation before quantization.

For block \(l\), standard calibration optimizes the learnable parameters
\(\boldsymbol{\theta}_l\) by matching the quantized block output
\(\tilde{\mathbf{h}}_l\) to the full-precision output \(\mathbf{h}_l\):
\begin{equation}
\mathcal{L}_{\mathrm{uni}}^{l}(\boldsymbol{\theta}_l)
=
\mathbb{E}_{x}
\left[
\frac{1}{|\mathcal{T}_l|C}
\sum_{t \in \mathcal{T}_l}
\sum_{c=1}^{C}
\bigl(
h_{l,t,c}(x)
-
\tilde{h}_{l,t,c}(x;\boldsymbol{\theta}_l)
\bigr)^2
\right],
\end{equation}
where \(\mathcal{T}_l\) denotes the tokens of block \(l\). This uniform loss
weights all channels equally, although errors in different channels can have
very different effects on downstream task predictions.

\paragraph{Task-aware Fisher aggregation.}
VGGT predicts multiple tasks, and each task head gives a Fisher score
\(F_k[l,c]\) for channel \(c\) in block \(l\). Since raw gradient magnitudes can
differ across task heads, we first normalize each task by its mean Fisher value:
\begin{equation}
\overline{F}_k
=
\frac{1}{LC}
\sum_{l'=1}^{L}
\sum_{c'=1}^{C}
F_k[l',c'] .
\end{equation}
We then combine the normalized scores with equal task weights:
\begin{equation}
s[l,c]
=
\sum_{k=1}^{K}
\frac{F_k[l,c]}{\overline{F}_k}.
\end{equation}
Here, \(L\) is the number of calibrated blocks and \(C\) is the hidden channel
dimension. This normalization prevents a task with larger raw gradients from
dominating the calibration weights.

\paragraph{Per-block normalization.}
Because calibration is performed block by block, we normalize the combined
score within each block:
\begin{equation}
w_l[c]
=
\frac{s[l,c]}
{\frac{1}{C}\sum_{c'=1}^{C}s[l,c']} .
\end{equation}
Thus, each block has mean weight \(1\), so the loss scale is comparable across
blocks. We also apply a small floor
\begin{equation}
w_l[c] \leftarrow \max\bigl(w_l[c], 0.01\bigr),
\end{equation}
which prevents low-Fisher channels from being completely ignored.

\paragraph{Fisher-guided calibration loss.}
\textbf{FGQ} replaces the uniform calibration loss with a Fisher-weighted objective:
\begin{equation}
\label{eq:cal}
\mathcal{L}_{\mathrm{FGQ}}^{l}(\boldsymbol{\theta}_l)
=
\mathbb{E}_{x}
\left[
\frac{1}{|\mathcal{T}_l|C}
\sum_{t \in \mathcal{T}_l}
\sum_{c=1}^{C}
w_l[c]
\bigl(
h_{l,t,c}(x)
-
\tilde{h}_{l,t,c}(x;\boldsymbol{\theta}_l)
\bigr)^2
\right].
\end{equation}
This objective allocates more calibration capacity to channels whose errors are
more likely to affect VGGT task outputs, while still preserving all channels
through the floor term.


\section{Evaluation}\label{sec:eval}
\subsection{Experimental Setup}


\paragraph{Evaluation settings.}

We primarily evaluate \textbf{FGQ} on the open-source VGGT-1.2B model~\citep{wang2025vggt}. 
We consider the main 3D vision tasks supported by VGGT and evaluate each task on standard benchmarks: 
(i) \textbf{camera pose estimation} on Co3Dv2~\citep{reizenstein2021common} and Re10K~\citep{zhou2018stereo}; 
(ii) \textbf{point map reconstruction} on 7-Scenes~\citep{shotton2013scene}, DTU~\citep{jensen2014large}, and ETH3D~\citep{schops2017multi}; 
and (iii) \textbf{depth estimation} on KITTI~\citep{geiger2013vision} for both monocular depth and video depth settings.

For the learnable affine quantization scheme, we implement FGQ on top of FlatQuant~\citep{sun2025flatquant}. 
FlatQuant is used as a strong and public quantization backbone, while FGQ elaborates the calibration objective. Detailed quantization and calibration settings are provided in Appendix~\ref{app:calibration_settings}. 
We report results under both \textbf{W4A8} and \textbf{W4A4} settings to evaluate FGQ under different quantization regimes. 
To further test generalization, we also extend \textbf{FGQ} to the \(\pi^3\) model~\citep{wang2025pi}. The result can be found in Appendix~\ref{app:pi3}.

\paragraph{Baselines.}
We compare \textbf{FGQ} against representative PTQ methods: (i) \textbf{RTN}: round-to-nearest quantization without calibration, serving as a lower bound; (ii) \textbf{QuaRot}~\citep{ashkboos2024quarot}: fixed Hadamard rotation for activation smoothing; (iii) \textbf{FlatQuant}~\citep{sun2025flatquant}: learnable affine transformations with uniform reconstruction loss, which serves as our base method; (iv) \textbf{QuantVGGT}~\citep{feng2025quantized}: a recent VGGT-specific PTQ method that addresses activation outliers from special tokens. For fair comparison, all methods use the same calibration set and quantization configuration.

\subsection{Main Results}
\begin{table}[t]
  \caption{Camera pose estimation results for VGGT on Co3Dv2~\citep{reizenstein2021common} and Re10K~\citep{zhou2018stereo}. AUC metrics measure the area under the cumulative error curve at different thresholds (higher is better). Best results in \textbf{bold}.}
  \label{tab:pose}
  \centering
  \resizebox{\textwidth}{!}{%
  \begin{tabular}{c|c|cccc|cccc}
    \toprule
    \multirow{2}{*}{Method} & \multirow{2}{*}{\shortstack{Bitwidth\\(W/A)}} & \multicolumn{4}{c|}{Co3Dv2} & \multicolumn{4}{c}{Re10K} \\
    \cmidrule(lr){3-6} \cmidrule(lr){7-10}
    & & AUC@30$\uparrow$ & AUC@15$\uparrow$ & AUC@5$\uparrow$ & AUC@3$\uparrow$ & AUC@30$\uparrow$ & AUC@15$\uparrow$ & AUC@5$\uparrow$ & AUC@3$\uparrow$ \\
    \midrule
    FP16 & 16/16 & 0.9731 & 0.9462 & 0.8462 & 0.7630 & 0.8667 & 0.7818 & 0.5803 & 0.4688 \\
    \midrule
    RTN & 4/8 & 0.9619 & 0.9238 & 0.7822 & 0.6691 & 0.8483 & 0.7514 & 0.5318 & 0.4158 \\
    QuaRot & 4/8 & 0.9672 & 0.9343 & 0.8098 & 0.7032 & 0.8480 & 0.7537 & 0.5307 & 0.4038 \\
    FlatQuant & 4/8 & 0.9674 & 0.9351 & 0.8051 & 0.7191 & 0.8481 & 0.7543 & 0.5262 & 0.3958 \\
    QuantVGGT & 4/8 & 0.9684 & 0.9363 & 0.8127 & 0.7117 & 0.8516 & 0.7587 & 0.5379 & 0.4125 \\
    \cellcolor{lightgray}\textbf{FGQ} & \cellcolor{lightgray}\textbf{4/8} & \cellcolor{lightgray}\textbf{0.9686} & \cellcolor{lightgray}\textbf{0.9371} & \cellcolor{lightgray}\textbf{0.8185} & \cellcolor{lightgray}\textbf{0.7230} & \cellcolor{lightgray}\textbf{0.8566} & \cellcolor{lightgray}\textbf{0.7651} & \cellcolor{lightgray}\textbf{0.5485} & \cellcolor{lightgray}\textbf{0.4318} \\
    \midrule
    RTN & 4/4 & 0.6547 & 0.3950 & 0.0474 & 0.0123 & 0.4334 & 0.2520 & 0.0590 & 0.0224 \\
    QuaRot & 4/4 & 0.8169 & 0.6471 & 0.2252 & 0.0904 & 0.5722 & 0.3952 & 0.1381 & 0.0663 \\
    FlatQuant & 4/4 & 0.9295 & 0.8648 & 0.6405 & 0.4691 & 0.7240 & 0.5849 & 0.3233 & 0.2169 \\
    QuantVGGT & 4/4 & 0.9387 & 0.8840 & 0.6973 & 0.5419 & 0.7703 & 0.6437 & 0.3845 & 0.2710 \\
    \cellcolor{lightgray}\textbf{FGQ} & \cellcolor{lightgray}\textbf{4/4} & \cellcolor{lightgray}\textbf{0.9425} & \cellcolor{lightgray}\textbf{0.8887} & \cellcolor{lightgray}\textbf{0.7025} & \cellcolor{lightgray}\textbf{0.5575} & \cellcolor{lightgray}\textbf{0.7704} & \cellcolor{lightgray}\textbf{0.6443} & \cellcolor{lightgray}\textbf{0.3898} & \cellcolor{lightgray}\textbf{0.2756} \\
    \bottomrule
  \end{tabular}%
  }

\end{table}

\begin{table}[t]
  \caption{Point map reconstruction results for VGGT on 7-Scenes~\citep{shotton2013scene} and ETH3D~\citep{schops2017multi}. Acc measures accuracy (higher is better), Comp measures completeness (lower is better), and N.C. measures normal consistency (higher is better). Best results in \textbf{bold}.}
  \label{tab:point}
  \centering
  \resizebox{\textwidth}{!}{%
  \begin{tabular}{c|c|cccccc|cccccc}
    \toprule
    \multirow{3}{*}{Method} & \multirow{3}{*}{\shortstack{Bitwidth\\(W/A)}} & \multicolumn{6}{c|}{7-Scenes} & \multicolumn{6}{c}{ETH3D} \\
    \cmidrule(lr){3-8} \cmidrule(lr){9-14}
    & & \multicolumn{2}{c}{Acc.$\downarrow$} & \multicolumn{2}{c}{Comp.$\downarrow$} & \multicolumn{2}{c|}{N.C.$\uparrow$} & \multicolumn{2}{c}{Acc.$\downarrow$} & \multicolumn{2}{c}{Comp.$\downarrow$} & \multicolumn{2}{c}{N.C.$\uparrow$} \\
    \cmidrule(lr){3-4} \cmidrule(lr){5-6} \cmidrule(lr){7-8} \cmidrule(lr){9-10} \cmidrule(lr){11-12} \cmidrule(lr){13-14}
    & & mean & med & mean & med & mean & med & mean & med & mean & med & mean & med \\
    \midrule
    FP16 & 16/16 & 0.044 & 0.024 & 0.056 & 0.033 & 0.733 & 0.846 & 0.263 & 0.167 & 0.288 & 0.167 & 0.846 & 0.947 \\
    \midrule
    RTN & 4/8 & 0.046 & 0.027 & 0.060 & 0.037 & 0.729 & 0.839 & 0.271 & 0.183 & 0.299 & 0.174 & 0.840 & 0.938 \\
    QuaRot & 4/8 & 0.047 & 0.027 & 0.056 & 0.033 & \textbf{0.736} & \textbf{0.848} & 0.262 & 0.174 & 0.272 & 0.161 & 0.846 & 0.951 \\
    FlatQuant & 4/8 & 0.045 & 0.025 & \textbf{0.055} & \textbf{0.031} & 0.732 & 0.845 & 0.262 & 0.169 & 0.287 & 0.168 & 0.845 & 0.943 \\
    QuantVGGT & 4/8 & 0.044 & 0.025 & 0.056 & 0.033 & 0.729 & 0.840 & 0.269 & 0.185 & 0.292 & 0.170 & 0.842 & 0.942 \\
    \rowcolor{lightgray}\textbf{FGQ} & \textbf{4/8} & \textbf{0.043} & \textbf{0.024} & \textbf{0.055} & \textbf{0.031} & \textbf{0.736} & \textbf{0.848} & \textbf{0.260} & \textbf{0.165} & \textbf{0.270} & \textbf{0.158} & \textbf{0.848} & \textbf{0.949} \\
    \midrule
    RTN & 4/4 & 0.146 & 0.103 & 0.134 & 0.086 & 0.600 & 0.655 & 0.944 & 0.843 & 1.339 & 0.920 & 0.603 & 0.656 \\
    QuaRot & 4/4 & 0.107 & 0.079 & 0.165 & 0.128 & 0.663 & 0.749 & 0.917 & 0.781 & 1.151 & 0.757 & 0.667 & 0.753 \\
    FlatQuant & 4/4 & 0.056 & 0.036 & 0.070 & 0.047 & 0.717 & 0.824 & 0.291 & 0.193 & 0.297 & 0.173 & 0.833 & 0.935 \\
    QuantVGGT & 4/4 & 0.053 & 0.032 & 0.085 & 0.059 & 0.719 & 0.828 & 0.312 & 0.206 & 0.305 & 0.180 & 0.832 & 0.937 \\
    \rowcolor{lightgray}\textbf{FGQ} & \textbf{4/4} & \textbf{0.048} & \textbf{0.031} & \textbf{0.059} & \textbf{0.036} & \textbf{0.723} & \textbf{0.832} & \textbf{0.275} & \textbf{0.177} & \textbf{0.279} & \textbf{0.164} & \textbf{0.834} & \textbf{0.940} \\
    \bottomrule
  \end{tabular}%
  }

\end{table}

\paragraph{Camera pose estimation.}
Table~\ref{tab:pose} evaluates camera pose estimation on Co3Dv2 and Re10K.
Under W4A8, most methods remain close to FP16, indicating that this task is
relatively robust to moderate activation quantization. In this setting, \textbf{FGQ}
still gives the best overall results on Re10K and competitive results on
Co3Dv2. In the more challenging
W4A4 setting, where activation quantization introduces a larger performance
drop, \textbf{FGQ} consistently performs on par with or better than existing
quantization baselines across datasets. Compared with the
dense reconstruction results below, the improvement on pose estimation is
modest, suggesting that this global prediction task is less sensitive to the
Fisher-guided channel selection than dense geometric outputs.

\paragraph{Point map reconstruction.}
Table~\ref{tab:point} and~\ref{tab:point_eth3d} report point map
reconstruction results on 7-Scenes, ETH3D, and DTU. \textbf{FGQ} shows a clearer
advantage on this task. 
Point maps are dense geometric outputs, so activation
quantization errors affect many spatial locations and can directly degrade
surface accuracy, completeness, and normal consistency. Under W4A8, \textbf{FGQ}
closely matches FP16 and obtains the strongest overall performance. Under
W4A4, the advantage becomes more pronounced. On 7-Scenes, \textbf{FGQ} reduces the mean
completeness error from 0.085 to 0.059 compared with QuantVGGT, while also
improving mean accuracy from 0.053 to 0.048. On ETH3D, \textbf{FGQ} improves mean
accuracy from 0.312 to 0.275 and mean completeness from 0.305 to 0.279. On
DTU, \textbf{FGQ} also gives the best completeness and normal consistency among W4A4
methods. These results indicate that \textbf{FGQ} better preserves the features needed
for dense 3D geometry. Qualitative comparisons are shown in Figure~\ref{fig:visual} and
Appendix~\ref{app:visual}. On 7-Scenes, \textbf{FGQ} (W4A4) preserves
finer structural details than QuantVGGT (W4A4), while on DTU the
QuantVGGT (W4A4) reconstruction appears visibly blurry and
\textbf{FGQ} (W4A4) remains sharp.

\paragraph{Depth estimation.}
\setlength{\intextsep}{0pt}
\setlength{\abovecaptionskip}{2pt}
\begin{wraptable}[17]{r}{0.65\textwidth}
  \caption{Point map reconstruction for VGGT on DTU~\citep{jensen2014large}. Best results in \textbf{bold}.}
  \label{tab:point_eth3d}
  \centering
  \scriptsize
  \begin{tabular}{c|c|cccccc}
    \toprule
    \multirow{3}{*}{Method} & \multirow{3}{*}{\shortstack{Bitwidth\\(W/A)}} & \multicolumn{6}{c}{DTU} \\
    \cmidrule(lr){3-8}
    & & \multicolumn{2}{c}{Acc.$\downarrow$} & \multicolumn{2}{c}{Comp.$\downarrow$} & \multicolumn{2}{c}{N.C.$\uparrow$} \\
    \cmidrule(lr){3-4} \cmidrule(lr){5-6} \cmidrule(lr){7-8}
    & & Mean & Med & Mean & Med & Mean & Med \\
    \midrule
    FP16 & 16 & 1.308 & 0.761 & 1.929 & 1.015 & 0.665 & 0.750 \\
    \midrule
    RTN & 4/8 & 1.363 & 0.786 & 1.904 & 0.976 & 0.669 & 0.753 \\
    QuaRot & 4/8 & 1.401 & 0.815 & 1.899 & 0.978 & 0.668 & 0.754 \\
    FlatQuant & 4/8 & 1.429 & 0.840 & 1.910 & 0.977 & 0.665 & 0.752 \\
    QuantVGGT & 4/8 & 1.292 & 0.752 & 1.944 & 1.007 & 0.667 & 0.753 \\
    \rowcolor{lightgray}\textbf{FGQ} & \textbf{4/8} & \textbf{1.287} & \textbf{0.742} & \textbf{1.881} & \textbf{0.965} & \textbf{0.675} & \textbf{0.762} \\
    \midrule
    RTN & 4/4 & 7.987 & 5.648 & 4.355 & 2.593 & 0.657 & 0.731 \\
    QuaRot & 4/4 & 3.470 & 2.107 & 2.145 & 1.077 & 0.665 & 0.750 \\
    FlatQuant & 4/4 & 1.478 & 0.829 & 1.895 & 0.982 & 0.667 & 0.754 \\
    QuantVGGT & 4/4 & 1.488 & 0.837 & 1.933 & 1.001 & \textbf{0.669} & 0.756 \\
    \rowcolor{lightgray}\textbf{FGQ} & \textbf{4/4} & \textbf{1.420} & \textbf{0.801} & \textbf{1.879} & \textbf{0.965} & \textbf{0.669} & \textbf{0.757} \\
    \bottomrule
  \end{tabular}
\end{wraptable}%
Table~\ref{tab:depth} reports depth estimation results on KITTI for both
monocular and video inputs. \textbf{FGQ} consistently outperforms the quantization
baselines in the W4A8 setting, achieving the best results across all metrics on
both evaluation protocols. The gains are especially clear on error metrics such
as AbsRel, SqRel, and RMSE, showing that Fisher-guided calibration better
preserves the dense depth structure under activation quantization. In the more
aggressive W4A4 setting, \textbf{FGQ} remains the strongest overall method. It achieves
the best AbsRel, SqRel, and threshold accuracy on KITTI Mono, and the best
overall error metrics on KITTI Video among the W4A4 methods. These results are
consistent with the point map reconstruction experiments: \textbf{FGQ} provides larger
benefits on dense geometric tasks than on global pose estimation, since dense
outputs are more sensitive to spatially accumulated quantization errors.
\begin{table}[t]
  \caption{Depth estimation results on KITTI (monocular and video depth). Lower is better for AbsRel, SqRel, and RMSE; higher is better for $\delta < 1.25$. Best results in \textbf{bold}.}
  \label{tab:depth}
  \centering
  \resizebox{\textwidth}{!}{%
  \begin{tabular}{c|c|cccc|cccc}
    \toprule
    \multirow{2}{*}{Method} & \multirow{2}{*}{Bitwidth (W/A)} & \multicolumn{4}{c|}{KITTI (Mono)} & \multicolumn{4}{c}{KITTI (Video)} \\
    \cmidrule(lr){3-6} \cmidrule(lr){7-10}
    & & AbsRel$\downarrow$ & SqRel$\downarrow$ & RMSE$\downarrow$ & $\delta<1.25$$\uparrow$ & AbsRel$\downarrow$ & SqRel$\downarrow$ & RMSE$\downarrow$ & $\delta<1.25$$\uparrow$ \\
    \midrule
    FP16 & 16/16 & 0.092 & 0.459 & 3.902 & 0.936 & 0.058 & 0.373 & 3.618 & 0.961 \\
    \midrule
    RTN       & 4/8 & 0.093 & 0.453 & 3.903 & 0.936 & 0.058 & 0.348 & 3.500 & 0.964 \\
    QuaRot    & 4/8 & 0.087 & 0.443 & 3.897 & 0.943 & 0.059 & 0.364 & 3.642 & 0.963 \\
    FlatQuant & 4/8 & 0.087 & 0.410 & 3.623 & 0.946 & 0.052 & 0.314 & 3.350 & 0.968 \\
    QuantVGGT & 4/8 & 0.091 & 0.462 & 3.931 & 0.936 & 0.054 & 0.339 & 3.450 & 0.967 \\
    \rowcolor{lightgray}\textbf{FGQ} & \textbf{4/8} & \textbf{0.086} & \textbf{0.396} & \textbf{3.566} & \textbf{0.947} & \textbf{0.051} & \textbf{0.292} & \textbf{3.240} & \textbf{0.970} \\
    \midrule
    RTN       & 4/4 & 0.148 & 0.728 & 5.159 & 0.823 & 0.164 & 0.985 & 5.615 & 0.748 \\
    QuaRot    & 4/4 & 0.117 & 0.487 & 4.083 & 0.903 & 0.126 & 0.713 & 4.896 & 0.838 \\
    FlatQuant & 4/4 & 0.084 & 0.399 & 3.650 & 0.948 & 0.062 & 0.339 & 3.488 & 0.960 \\
    QuantVGGT & 4/4 & 0.088 & 0.446 & 3.842 & 0.938 & 0.058 & 0.336 & 3.486 & 0.963 \\
    \rowcolor{lightgray}\textbf{FGQ} & \textbf{4/4} & \textbf{0.079} & \textbf{0.389} & \textbf{3.642} & \textbf{0.951} & \textbf{0.056} & \textbf{0.313} & \textbf{3.384} & \textbf{0.965} \\
    \bottomrule
  \end{tabular}%
  }
\end{table}

\subsection{Ablation study}

We study the effect of different task losses in the empirical Fisher
estimation. All variants use the same quantization setting, calibration
data, optimization steps, and evaluation protocol. The plain baseline
removes Fisher weighting and uses an unweighted calibration objective.
Co3Dv2 mainly evaluates camera pose accuracy, while DTU evaluates
point-map reconstruction. As shown in Table~\ref{tab:ablation_fisher},
Fisher weighting improves both benchmarks. On Co3Dv2, adding point and
depth Fisher improves AUC@15 from \(0.8825\) to \(0.8875\), while further
adding camera Fisher gives the largest gain and reaches \(0.8952\). On
DTU, adding point Fisher to the camera-depth Fisher gives the best Acc.
mean, reducing it from \(1.458\) to \(1.446\). This shows that the Fisher
term is most useful when it includes the task aligned with the evaluation
metric. At the same time, the gains from multi-task Fisher suggest that
different task losses provide complementary channel sensitivity signals.

\begin{table}[t]
\centering
\caption{Ablation study on the Fisher Information design.}
\label{tab:ablation_fisher}
\resizebox{.8\linewidth}{!}{%
\begin{tabular}{l|c||l|c}
    \toprule
    Fisher Info. & Co3Dv2 AUC@15 $\uparrow$ & Fisher Info. & DTU Acc. mean $\downarrow$ \\
    \midrule
    plain (w/o Fisher)     & 0.8825          & plain (w/o Fisher)     & 1.537 \\
    point                  & 0.8868          & camera                 & 1.465 \\
    point + depth          & 0.8875          & camera + depth         & 1.458 \\
    point + depth + camera & \textbf{0.8952} & camera + depth + point & \textbf{1.446} \\
    \bottomrule
\end{tabular}%
}

\end{table}

\subsection{Efficiency Analysis}\label{subsec: efficiency}
\begin{wrapfigure}[15]{r}{0.4\textwidth}
    \centering
    \includegraphics[width=0.38\textwidth]{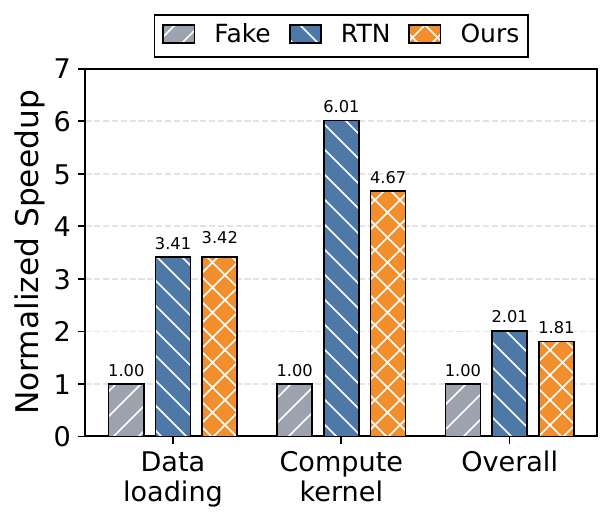}
    \caption{Normalized Speedup over the same W4A4 quantization settings.}
    \label{fig:efficiency}
\end{wrapfigure}%
We evaluate the runtime efficiency of our W4A4 quantized model on a single NVIDIA A100 40G GPU.
To measure the benefit of real low-bit
execution, we use fake quantization as the baseline and report speedups
from the real quantized implementation. Our implementation builds on the
Triton kernel from FlatQuant~\citep{sun2025flatquant} and tailors it to our design. We decompose
the latency into parameter loading and compute kernel time, where
the compute kernel includes the affine transformation, quantization, and
quantized linear computation. For parameter loading, RTN and our method
achieve similar speedups, \(3.41\times\) and \(3.42\times\), since both
benefit from 4-bit weight and activation storage. For the compute kernel,
RTN reaches \(6.01\times\), while our method reaches \(4.67\times\). The
gap comes from the extra affine transformation used by our method. At the
full block level, RTN obtains \(2.01\times\) speedup and our method still
achieves \(1.81\times\) speedup. These results show that our method keeps
most of the real quantization efficiency while adding only moderate
runtime overhead for the accuracy-preserving affine transformation.

\section{Conclusion}

We presented Fisher-Guided Quantization (\textbf{FGQ}), a post-training
quantization method for feed-forward 3D reconstruction models. Our main
observation is that different transformer blocks and
hidden channels in the model contribute differently to each task. 
As a result, uniformly treating all outputs during calibration can therefore preserve
insensitive tasks while harming the tasks that are more sensitive to
quantization noise. \textbf{FGQ} addresses this issue by
using diagonal Fisher information to estimate task-aware block and channel
sensitivities, and by incorporating these sensitivities into the learnable
affine transformation during calibration. Extensive experiments show that 
\textbf{FGQ} consistently outperforms existing PTQ baselines under 4-bit quantization,
with the largest gains appearing on point map reconstruction, where prior PTQ methods suffer the most severe degradation.

\begin{ack}
This work is supported by AI Chip Center for
Emerging Smart Systems (ACCESS), Hong Kong SAR and
Collaborative Research Fund (UGC CRF) C5032-23G.
\end{ack}




\bibliographystyle{plainnat}
\bibliography{references}

\newpage
\appendix

\section{Proof of Proposition~\ref{prop:fisher_hessian}}
\label{app:proof}

We prove the identity for a continuous label space. The same argument
applies to discrete labels by replacing integrals with sums. Fix an
input \(x\). Since \(p_z(y\mid x)\) is a normalized conditional
distribution,
\begin{equation}
\int p_z(y\mid x)\,dy = 1.
\label{eq:normalization}
\end{equation}
Under the stated regularity conditions, we may differentiate
Eq.~\eqref{eq:normalization} with respect to \(z\). Differentiating once
gives
\begin{equation}
0
=
\int \nabla_z p_z(y\mid x)\,dy
=
\int p_z(y\mid x)\nabla_z \log p_z(y\mid x)\,dy.
\label{eq:score_identity}
\end{equation}
Thus the expected score is zero:
\begin{equation}
\mathbb{E}_{y\sim p_z(\cdot\mid x)}
\!\left[
\nabla_z \log p_z(y\mid x)
\right]
=0.
\end{equation}

Differentiating Eq.~\eqref{eq:score_identity} once more gives
\begin{align}
0
&=
\int \nabla_z
\left[
p_z(y\mid x)\nabla_z \log p_z(y\mid x)
\right]dy \nonumber \\
&=
\int p_z(y\mid x)
\left[
\nabla_z^2 \log p_z(y\mid x)
+
\nabla_z \log p_z(y\mid x)
\nabla_z \log p_z(y\mid x)^{\!\top}
\right]dy.
\label{eq:second_score_identity}
\end{align}
Equivalently,
\begin{equation}
\mathbb{E}_{y\sim p_z(\cdot\mid x)}
\!\left[
-\nabla_z^2 \log p_z(y\mid x)
\right]
=
\mathbb{E}_{y\sim p_z(\cdot\mid x)}
\!\left[
\nabla_z \log p_z(y\mid x)
\nabla_z \log p_z(y\mid x)^{\!\top}
\right].
\label{eq:bartlett_loglik}
\end{equation}

Now define the negative log-likelihood loss
\begin{equation}
\ell(z;x,y)=-\log p_z(y\mid x).
\end{equation}
Then
\begin{equation}
\nabla_z^2 \ell(z;x,y)
=
-\nabla_z^2 \log p_z(y\mid x),
\qquad
\nabla_z \ell(z;x,y)
=
-\nabla_z \log p_z(y\mid x).
\end{equation}
Substituting these relations into Eq.~\eqref{eq:bartlett_loglik} yields
\begin{equation}
\mathbb{E}_{y\sim p_z(\cdot\mid x)}
\!\left[
\nabla_z^2 \ell(z;x,y)
\right]
=
\mathbb{E}_{y\sim p_z(\cdot\mid x)}
\!\left[
\nabla_z \ell(z;x,y)
\nabla_z \ell(z;x,y)^{\!\top}
\right].
\label{eq:conditional_identity}
\end{equation}

Evaluating Eq.~\eqref{eq:conditional_identity} at \(z=z^\star\) and
using the assumption \(q(y\mid x)=p_{z^\star}(y\mid x)\), we obtain
\begin{equation}
\mathbb{E}_{y\sim q(\cdot\mid x)}
\!\left[
\nabla_z^2 \ell(z^\star;x,y)
\right]
=
\mathbb{E}_{y\sim q(\cdot\mid x)}
\!\left[
\nabla_z \ell(z^\star;x,y)
\nabla_z \ell(z^\star;x,y)^{\!\top}
\right].
\end{equation}
Finally, taking expectation over \(x\sim q(x)\) gives
\begin{equation}
\mathbb{E}_{x\sim q(x),\,y\sim q(y\mid x)}
\!\left[
\nabla_z^2 \ell(z^\star;x,y)
\right]
=
\mathbb{E}_{x\sim q(x),\,y\sim q(y\mid x)}
\!\left[
\nabla_z \ell(z^\star;x,y)
\nabla_z \ell(z^\star;x,y)^{\!\top}
\right],
\end{equation}
which proves Proposition~\ref{prop:fisher_hessian}.
\qed

\section{Quantization and Calibration Settings}
\label{app:calibration_settings}

\paragraph{Implementation based on FlatQuant.}
FGQ is implemented on top of FlatQuant~\citep{sun2025flatquant}. 
We use the same learnable affine transformation modules, learnable clipping parameters, and quantizer placement as FlatQuant. 
All pretrained model parameters are frozen before calibration. 
Calibration is performed block by block, and only the calibration parameters in the current block are optimized. 
Specifically, the trainable parameters are the FlatQuant transformation parameters, including the transformation matrices and diagonal scales, together with the learnable weight-clipping and activation-clipping factors. 
The original linear weights and biases, normalization layers, patch embedding/DINOv2 backbone, prediction heads, and all non-current blocks are kept frozen. 
This setup isolates the effect of replacing the standard FlatQuant reconstruction loss with the proposed Fisher-guided calibration loss.

\paragraph{Quantization setting.}
For the default W4A4 setting, both weights and activations are quantized to \(4\) bits. 
Weight and activation quantization are symmetric. 
Weights are quantized per output channel, while activations are quantized per token without grouping. 
Learnable weight clipping and learnable activation clipping are enabled. 
Quantization is applied to the attention projections, including \(Q\), \(K\), \(V\), and output projections, as well as the MLP linear layers in the VGGT aggregator frame and global blocks. 
The DINOv2/patch embedding module, camera/depth/point/track heads, normalization parameters, and special tokens are not quantized.

\paragraph{Tasks and metrics.}
VGGT jointly predicts three tasks. For \textbf{depth 
estimation}, we report absolute relative error (AbsRel) 
and $\delta_1$ accuracy (percentage of pixels with 
relative error $< 1.25$). For \textbf{camera pose 
estimation}, we report the Area Under the Curve (AUC) 
of pose accuracy across multiple thresholds. 
For \textbf{point map reconstruction}, we report Accuracy 
(Acc.), Completeness (Comp.), and Normal Consistency 
(N.C.), with both mean and median values.

\paragraph{Calibration data.}
We use Co3Dv2~\citep{reizenstein2021common} as the calibration data source. 
For calibration, we sample \(64\) clips, with \(4\) frames per clip. 
The batch size is \(2\) clips, corresponding to \(8\) frames per mini-batch. 
The default random seed is \(42\). 
The sampled calibration clips are loaded once and then reused for all blocks and all calibration epochs. 
The implementation samples directly from the raw Co3Dv2 directory by first sampling a category, then a sequence, and then \(4\) images from the sequence image folder. 
The current sampler does not use annotation-based quality filters. 
Frames are randomly sampled from the available images in each sequence, and no explicit temporal stride or temporal ordering constraint is imposed.


\paragraph{Optimization hyperparameters.}
Calibration uses AdamW. 
Unless overridden by parameter groups, AdamW uses the PyTorch default hyperparameters: \(\beta_1=0.9\), \(\beta_2=0.999\), \(\epsilon=10^{-8}\), and weight decay \(0.01\). 
No gradient clipping is used. 
Mixed precision is enabled by default using CUDA bfloat16 autocast, without a gradient scaler. 
The base learning rate for the FlatQuant transformation matrices and diagonal scales is \(10^{-2}\). 
The learning rate for the learnable weight-clipping and activation-clipping factors is \(10\) times larger, i.e., \(10^{-1}\) in the default W4A4 setting.

The learning rate is decayed with a cosine schedule. 
The scheduler is stepped once per optimizer step. 
For each block, the number of calibration epochs is \(15\). 
With \(64\) clips and a batch size of \(2\) clips, this gives \(32\) mini-batches per epoch and \(480\) optimizer steps per block. 
VGGT contains \(48\) calibrated aggregator blocks, so the full W4A4 calibration uses \(48 \times 480 = 23040\) optimizer steps. 
The cosine schedule uses
\[
T_{\max} = \texttt{epochs} \times \left(\texttt{nsamples} // \texttt{cali\_bsz}\right),
\]
with minimum learning rate \(\eta_{\min}=10^{-5}\) for the default setting. 

\paragraph{Model mode during calibration.}
The model is kept in evaluation mode during calibration. 
Thus, dropout and stochastic-depth layers are disabled, and batch-normalization layers, if present, use evaluation behavior. 
Gradients are enabled only for the current block's calibration parameters. 
For each block, the full-precision output of that block is first used as the reconstruction target, and the quantized block output is then optimized to match this target.

\paragraph{Fisher information estimation.}
For FGQ, the Fisher information is estimated once before calibration and then kept fixed throughout calibration. 
We use the same Co3Dv2 sampling configuration as calibration: \(64\) clips, \(4\) frames per clip, image size \(518\), and random seed \(42\). 
The Fisher-estimation clips and calibration clips are generated independently by the same deterministic sampling procedure, rather than being explicitly saved and reused by identity.

Our implementation uses a diagonal activation Fisher approximation rather than a parameter Fisher. 
Hooks are registered on the outputs of the VGGT aggregator frame and global blocks. 
For VGGT, this gives \(48\) calibrated blocks. 
We estimate Fisher information for three output tasks: camera, depth, and world points, corresponding to \(\texttt{pose\_enc}\), \(\texttt{depth}\), and \(\texttt{world\_points}\). 
For each task, we use the sum of the task output as the scalar objective and accumulate squared gradients with respect to the hooked block activations. 
For task \(k\), block \(\ell\), and channel \(c\), the raw Fisher estimate is
\[
F_{k,\ell,c}
=
\frac{1}{N}
\sum_{n=1}^{N}
\sum_{t}
\left(
\frac{\partial \mathcal{L}_{k}}
{\partial h_{\ell,t,c}^{(n)}}
\right)^2,
\]
where \(h_{\ell,t,c}^{(n)}\) is the activation of channel \(c\) at token \(t\) in block \(\ell\) for sample \(n\), and \(\mathcal{L}_{k}\) is the scalar objective for task \(k\). 
The Fisher tensor therefore has shape \(3 \times 48 \times 1024\), corresponding to tasks, blocks, and channels. 
Forward passes for Fisher estimation use CUDA bfloat16 autocast, while gradients are cast to float32 before accumulation. 
The saved Fisher tensor is stored in float32.

Before being used in FGQ calibration, the raw Fisher tensor is normalized. 
First, each task Fisher is divided by its global mean. 
The normalized task Fishers are then summed with equal weight. 
Finally, the resulting per-block Fisher weights are normalized so that each block has mean weight \(1\), and the minimum weight is clamped to \(0.01\). 
No clipping or damping is applied to the raw Fisher values before this loading-time normalization.

\paragraph{FGQ calibration loss.}
FlatQuant uses an unweighted block-output reconstruction loss,
\[
\mathcal{L}_{\mathrm{FlatQuant}}
=
\operatorname{MSE}
\left(
\mathbf{y}_{\ell}^{q},
\mathbf{y}_{\ell}^{\mathrm{fp}}
\right),
\]
where \(\mathbf{y}_{\ell}^{q}\) and \(\mathbf{y}_{\ell}^{\mathrm{fp}}\) denote the quantized and full-precision outputs of the current block \(\ell\), respectively. 
FGQ replaces this loss with a Fisher-weighted reconstruction loss,
\[
\mathcal{L}_{\mathrm{FGQ}}
=
\frac{1}{|\mathcal{T}|C}
\sum_{t \in \mathcal{T}}
\sum_{c=1}^{C}
w_{\ell,c}
\left(
y_{\ell,t,c}^{q}
-
y_{\ell,t,c}^{\mathrm{fp}}
\right)^2,
\]
where \(w_{\ell,c}\) is the fixed Fisher-derived weight for channel \(c\) in block \(\ell\), \(\mathcal{T}\) is the set of tokens, and \(C\) is the channel dimension. 
Calibration proceeds block by block in the order
\[
\texttt{frame\_0}, \texttt{global\_0}, \ldots, \texttt{frame\_23}, \texttt{global\_23}.
\]
Before backpropagation, the scalar loss is divided by its detached value. 
This normalization preserves the gradient direction while stabilizing the gradient scale.


\section{Calibration Overhead}\label{app:overhead}
All calibration experiments are conducted on a single NVIDIA A100 40GB GPU.
In a representative W4A4 run, FlatQuant takes \(52.53\) minutes for calibration, while FGQ takes \(55.54\) minutes.
The corresponding peak GPU memory usage is \(3.54\) GB for FlatQuant and \(3.62\) GB for FGQ.
Compared with FlatQuant, FGQ increases the calibration time by \(3.01\) minutes, corresponding to a relative overhead of approximately \(5.73\%\), and increases the peak GPU memory usage by only \(0.08\) GB, corresponding to approximately \(2.26\%\).
Because Fisher estimation uses only \(64\) clips with \(4\) frames per clip and is performed once, its cost remains small compared with model training.
The detailed calibration memory usage and computation time are given in Table~\ref{tab:calibration}.

\begin{table}[t]
  \caption{Calibration overhead and end-task performance comparison. Numbers in \textcolor{red}{\small red} indicate the change of FGQ relative to the Baseline PTQ (FlatQuant).}
  \label{tab:calibration}
  \centering
  \resizebox{\textwidth}{!}{%
  \begin{tabular}{l|ll|ll}
    \toprule
    \multirow{2}{*}{Method} & \multicolumn{2}{c|}{Calibration Overhead} & \multicolumn{2}{c}{Performance} \\
    \cmidrule(lr){2-3} \cmidrule(lr){4-5}
    & GPU memory (GB) & GPU time (hours) & AUC@15 (Co3Dv2)$\uparrow$ & DTU Acc. mean$\downarrow$ \\
    \midrule
    FP16                       & --   & --   & 0.9462     & 1.308    \\
    \midrule
    Baseline PTQ (FlatQuant)   & 3.54 & 0.87 & 0.8825 & 1.537 \\
    \rowcolor{lightgray}\textbf{FGQ (Ours)}
      & \textbf{3.62}\textsubscript{\textcolor{red}{\tiny +0.08}}
      & \textbf{0.92}\textsubscript{\textcolor{red}{\tiny +0.05}}
      & \textbf{0.8952}\textsubscript{\textcolor{red}{\tiny +0.0127}}
      & \textbf{1.446}\textsubscript{\textcolor{red}{\tiny -0.091}} \\
    \bottomrule
  \end{tabular}%
  }
\end{table}
\section{FGQ Evaluation Result on $\pi^3$}
\label{app:pi3}
\begin{table}[t]
  \caption{$\pi^3$~\citep{wang2025pi} evaluation results. Top: camera pose estimation on Co3Dv2~\citep{reizenstein2021common} and Re10K~\citep{zhou2018stereo} (AUC, higher is better). Bottom: point map reconstruction on DTU~\citep{jensen2014large} and ETH3D~\citep{schops2017multi} (Acc and Comp lower is better; N.C. higher is better). Best results in \textbf{bold}.}
  \label{tab:pi3}
  \centering

  \subfloat{%
  \resizebox{\textwidth}{!}{%
  \begin{tabular}{c|c|cccc|cccc}
    \toprule
    \multirow{2}{*}{Method} & \multirow{2}{*}{\shortstack{Bitwidth\\(W/A)}} & \multicolumn{4}{c|}{Co3Dv2} & \multicolumn{4}{c}{Re10K} \\
    \cmidrule(lr){3-6} \cmidrule(lr){7-10}
    & & AUC@30$\uparrow$ & AUC@15$\uparrow$ & AUC@5$\uparrow$ & AUC@3$\uparrow$ & AUC@30$\uparrow$ & AUC@15$\uparrow$ & AUC@5$\uparrow$ & AUC@3$\uparrow$ \\
    \midrule
    FP16      & 16/16 & 0.9441 & 0.8882 & 0.6811 & 0.5140 & 0.8907 & 0.8212 & 0.6511 & 0.5523 \\
    \midrule
    RTN       & 4/4   & 0.7684 & 0.5491 & 0.0954 & 0.0246 & 0.5662 & 0.4040 & 0.1538 & 0.0740 \\
    FlatQuant & 4/4   & 0.9119 & 0.8240 & 0.5022 & 0.2686 & 0.8298 & 0.7317 & 0.5204 & 0.4142 \\
    \rowcolor{lightgray}\textbf{FGQ} & \textbf{4/4} & \textbf{0.9394} & \textbf{0.8795} & \textbf{0.6711} & \textbf{0.5032} & \textbf{0.8436} & \textbf{0.7502} & \textbf{0.5517} & \textbf{0.4461} \\
    \bottomrule
  \end{tabular}%
  }}

  \vspace{0.6em}

  \subfloat{%
  \resizebox{\textwidth}{!}{%
  \begin{tabular}{c|c|cccccc|cccccc}
    \toprule
    \multirow{3}{*}{Method} & \multirow{3}{*}{\shortstack{Bitwidth\\(W/A)}} & \multicolumn{6}{c|}{DTU} & \multicolumn{6}{c}{ETH3D} \\
    \cmidrule(lr){3-8} \cmidrule(lr){9-14}
    & & \multicolumn{2}{c}{Acc.$\downarrow$} & \multicolumn{2}{c}{Comp.$\downarrow$} & \multicolumn{2}{c|}{N.C.$\uparrow$} & \multicolumn{2}{c}{Acc.$\downarrow$} & \multicolumn{2}{c}{Comp.$\downarrow$} & \multicolumn{2}{c}{N.C.$\uparrow$} \\
    \cmidrule(lr){3-4} \cmidrule(lr){5-6} \cmidrule(lr){7-8} \cmidrule(lr){9-10} \cmidrule(lr){11-12} \cmidrule(lr){13-14}
    & & mean & med & mean & med & mean & med & mean & med & mean & med & mean & med \\
    \midrule
    FP16      & 16/16 & 1.138 & 0.606 & 1.926 & 0.624 & 0.662 & 0.747 & 0.188 & 0.123 & 0.211 & 0.128 & 0.884 & 0.971 \\
    \midrule
    RTN       & 4/4   & 7.744 & 5.814 & 19.290 & 17.076 & 0.585 & 0.630 & 1.353 & 1.263 & 2.030 & 1.647 & 0.589 & 0.628 \\
    FlatQuant & 4/4   & 1.396 & 0.745 & 1.917 & 0.622 & 0.662 & 0.746 & 0.197 & 0.134 & 0.217 & 0.145 & 0.870 & 0.959 \\
    \rowcolor{lightgray}\textbf{FGQ} & \textbf{4/4} & \textbf{1.358} & \textbf{0.726} & \textbf{1.878} & \textbf{0.599} & \textbf{0.663} & \textbf{0.747} & \textbf{0.193} & \textbf{0.130} & \textbf{0.215} & \textbf{0.131} & \textbf{0.874} & \textbf{0.961} \\
    \bottomrule
  \end{tabular}%
  }}
\end{table}

We further extend our method to $\pi^3$~\citep{wang2025pi}, the most recent feed-forward 3D reconstruction model. Table~\ref{tab:pi3} summarizes FGQ's performance on $\pi^3$.

On camera pose estimation, RTN collapses under 4/4 quantization, with AUC@3 on Co3Dv2 dropping from 0.5140 (FP16) to 0.0246, showing that $\pi^3$ is highly sensitive to quantization error. FlatQuant recovers a large portion of this gap but still suffers noticeable degradation at tight thresholds such as AUC@5 and AUC@3. FGQ consistently improves over FlatQuant on every threshold of both Co3Dv2 and Re10K, with the largest gains observed at tighter thresholds. For example, on Co3Dv2 FGQ raises AUC@3 from 0.2686 (FlatQuant) to 0.5032, approaching the FP16 result of 0.5140.

For point map reconstruction, FGQ outperforms FlatQuant on all metrics across DTU and ETH3D, and its results closely track the FP16 baseline (e.g., DTU Acc. mean 1.358 vs. FP16's 1.138, ETH3D N.C. mean 0.874 vs. FP16's 0.884). Overall, these results demonstrate that FGQ preserves $\pi^3$'s 3D geometric reasoning quality under aggressive 4-bit quantization while incurring minimal calibration overhead.

\section{Limitations}
\label{app:limitations}

FGQ has several limitations that we leave for future work. First, our experiments focus on standard INT4 weight-activation quantization. We have not yet studied how Fisher-guided calibration interacts with newer low-precision formats, such as MXFP4 or other floating-point 4-bit variants. Since these formats have different dynamic ranges and rounding behavior, they may benefit from different calibration schedules or Fisher weighting strategies.

Second, FGQ is evaluated with a fixed calibration protocol. We use the same calibration data scale and preprocessing across experiments to isolate the effect of Fisher-guided weighting. We do not attempt to optimize the calibration set construction, such as the category balance, scene diversity, or view distribution. Studying how to choose the most informative calibration samples for feed-forward 3D reconstruction is an interesting direction, but is orthogonal to the calibration objective proposed in this work.

\section{Impact Statement}\label{app:impact}
This work advances the practical deployment of feed-forward 3D reconstruction models by substantially reducing their memory and computational overhead through post-training quantization, enabling on-device 3D perception for applications such as augmented and virtual reality, robotics, autonomous driving, and assistive technologies. By lowering hardware requirements, FGQ helps democratize access to advanced 3D vision capabilities for researchers and developers with limited resources, while also reducing the energy consumption and carbon footprint associated with running billion-parameter models.

\section{More Visualization Results}\label{app:visual}

\begin{figure}
    \centering
    \includegraphics[width=\linewidth]{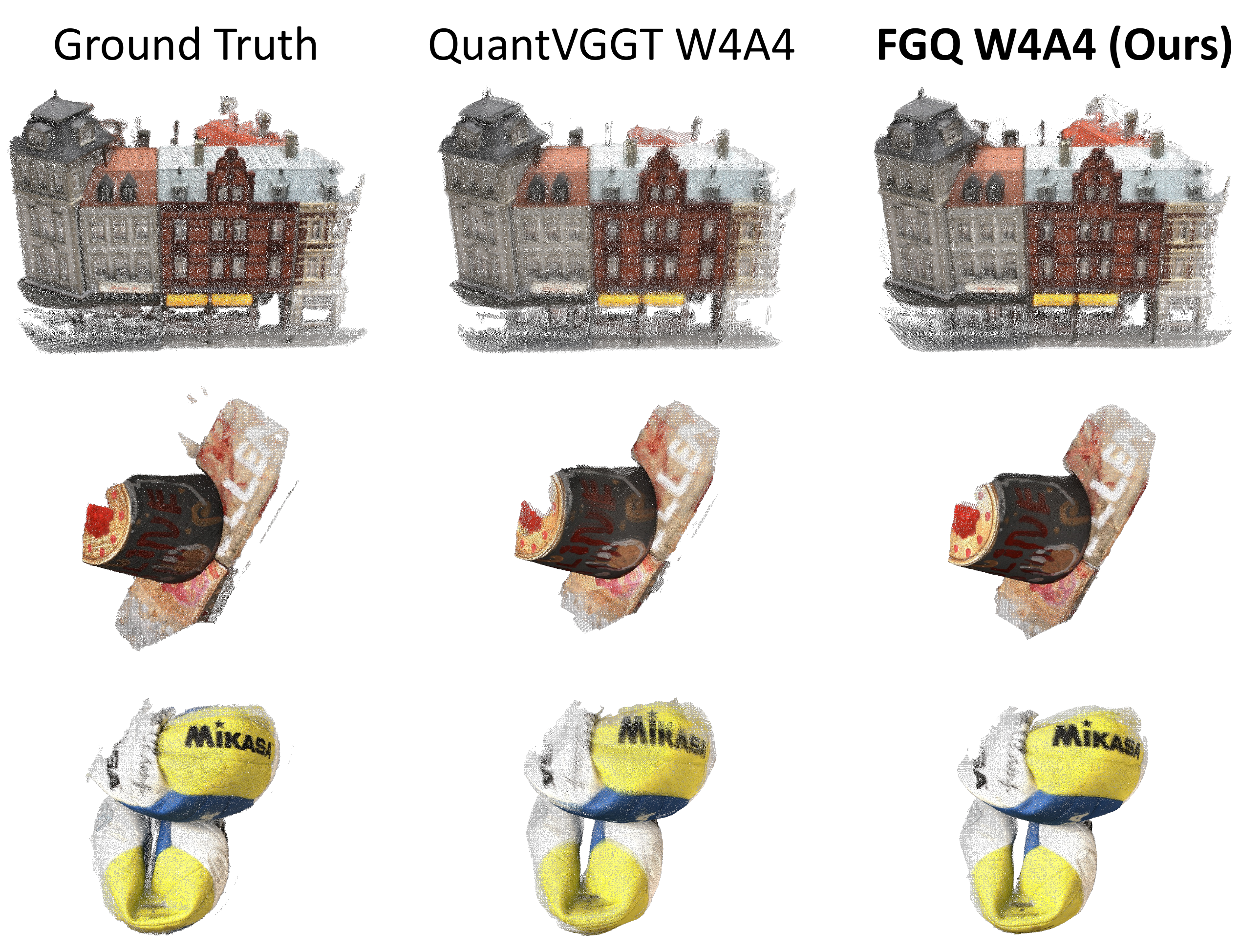}
    \caption{Visual comparison results for VGGT on DTU~\citep{jensen2014large} for Ground Truth, QuantVGGT~\citep{feng2025quantized} W4A4, and FGQ W4A4.}
    \label{fig:DTU}
\end{figure}

\begin{figure}
    \centering
    \includegraphics[width=\linewidth]{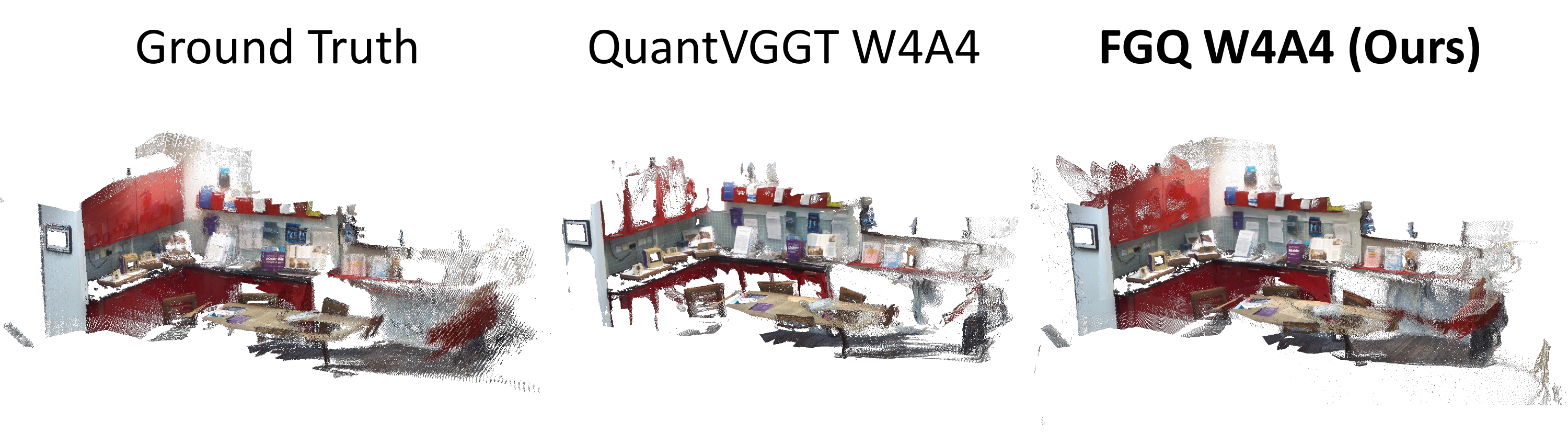}
    \caption{Visual comparison results for VGGT on 7-Scenes~\citep{shotton2013scene} for Ground Truth, QuantVGGT~\citep{feng2025quantized} W4A4, and FGQ W4A4.}
    \label{fig:7scenes}
\end{figure}

Here we exhibit more visualization result for VGGT~\citep{wang2025vggt} on DTU~\citep{jensen2014large} (Fig.~\ref{fig:DTU}) and 7-Scenes~\citep{shotton2013scene} (Fig.~\ref{fig:7scenes}). 

\begin{figure}
    \centering
    \includegraphics[width=\linewidth]{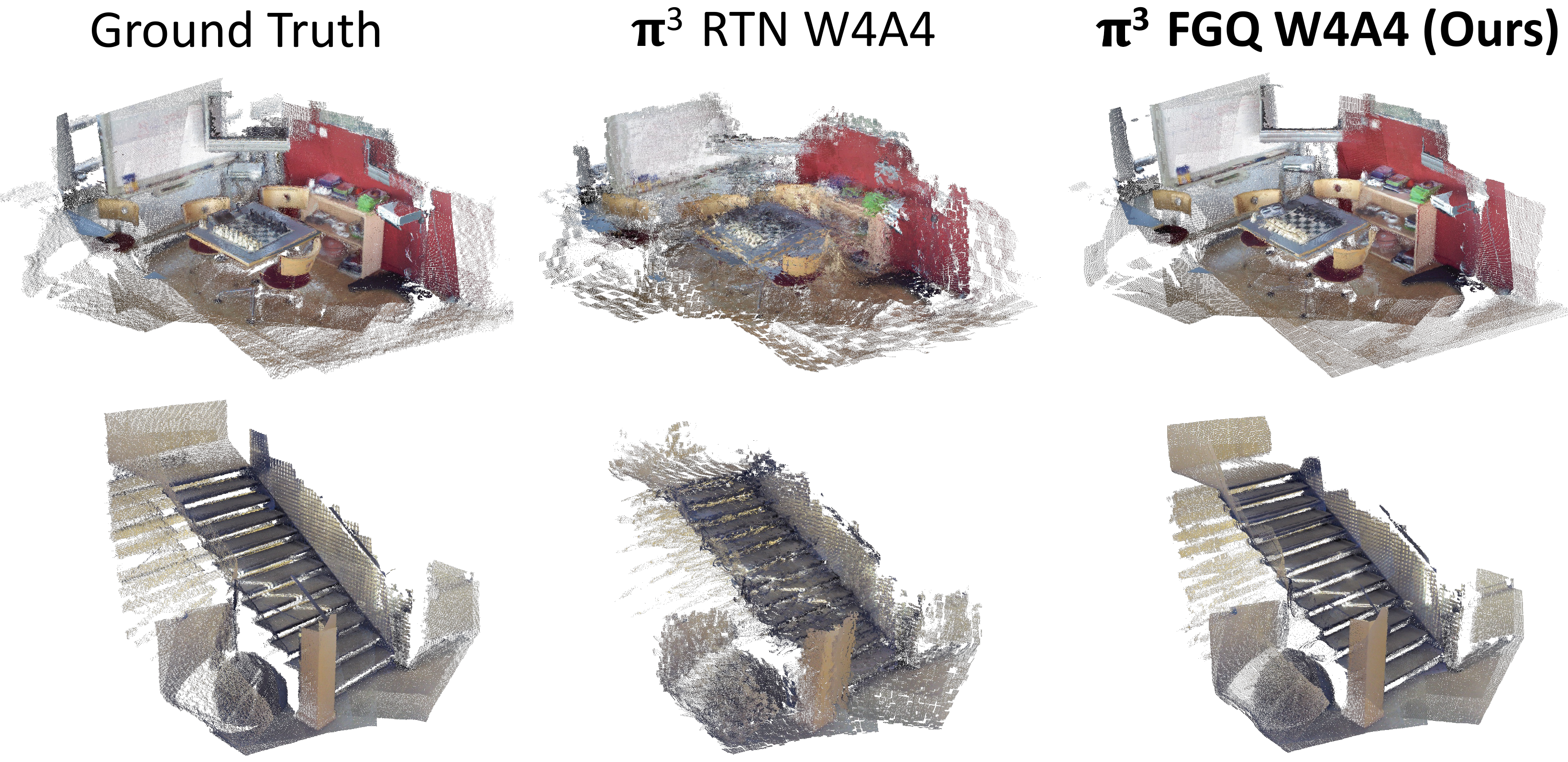}
    \caption{Visual comparison results for $\pi^3$ on 7-Scenes~\citep{shotton2013scene} for Ground Truth, RTN W4A4, and FGQ W4A4.}
    \label{fig:pi3_vis}
\end{figure}

Moreover, we provide the visualization result for $\pi^3$~\citep{wang2025pi} on 7-Scenes~\citep{shotton2013scene} in Fig.~\ref{fig:pi3_vis} to show FGQ's improvement over RTN. 



\end{document}